\documentclass{article}
\usepackage{cite}

   \usepackage[pdftex]{graphicx}
   \graphicspath{{./}{../}}
\usepackage{subcaption}

\usepackage{amsmath}
\usepackage{amssymb}
\usepackage{array}

\newtheorem{theorem}{Theorem}[section]
\newtheorem{corollary}{Corollary}[theorem]

\begin{document}
%
\title{Goldilocks Neural Networks}
%
%
%

\author{J.~Rosenzweig\thanks{Contact: jan.rosenzweig@brancherose.com}, Z.~Cvetkovi\' c, I.~Rosenzweig}

\maketitle

\begin{abstract}
We introduce the new "Goldilocks" class of activation functions, which non-linearly deform the input signal only locally when the input signal is in the appropriate range. The small local deformation of the signal enables better understanding of how and why the signal is transformed through the layers.  Numerical results on CIFAR-10 and CIFAR-100 data sets show that Goldilocks networks perform better than, or comparably to SELU and RELU, while introducing tractability of data deformation through the layers.
\end{abstract}


%

\section{Introduction}
%
%
%
%
Training deep neural networks is an important problem which is still far from solved. At the core of the problem is our still relatively poor understanding of what happens under the hood of a deep neural network. Practically, this translates to a wide variety of deep network architectures and activation functions used in them.

They all, however, suffer from the same problem when it comes to interpretability. It is next to impossible to understand how and why even a single layer network performs a simple classification task, and this probelm only increases with the size and the depth of the network.

Activation functions stem from Cybenko's seminal 1989 paper \cite{cybenko}, which proved that sigmoidal functions are universal approximators. This gave rise to a number of sigmoidal activation functions, including the sigmoid, tanh, arctan, binary step, Elliott sign \cite{elliott}, SoftSign \cite{quadratic} \cite{training}, SQNL \cite{sqnl}, soft clipping \cite{phase} and many others. 

Sigmoidal activations were useful in the early days of neural networks, but the most serious problem that they suffered from was vanishing gradients. Sigmoidal functions could easily become locked deep in either the activated or the inactivated state where the gradient of the activation function vanishes, and those neurons would became unreachable.

A partial solution to the vanishing gradient problem was found by introducing a linear tail, leading to a number of new activation functions such as the RELU \cite{relu}, which remains the most popular activation function in this class due to its simplicity, even though it still suffers from the vanishing gradient in the inactive state. The vanishing gradient has been fixed in a number of variations such as the leaky RELU \cite{leaky}, ISRLU \cite{isrlu}, BRELU \cite{bipolar}, PRELU \cite{prelu}, RRELU \cite{rrelu}, ELU \cite{elu}, SRELU \cite{sshape}, adaptive piecewise linear \cite{learning},  SoftPlus \cite{softplus}, sigmoid Linear Unit \cite{gelu}  \cite{sigmoidw} \cite{search}, SoftExponential \cite{continuum} and many others. 

There is also a number of "left field" activation functions such as linear activations, radial functions and the Gaussian, Fourier-series based activations \cite{fourier}, and even a recent study by Chen et al \cite{ode} where the entire shape of the activation function is the result of training, and not just its parameters.

The most significant recent development in the study of activation functions was the SELU activation of  Klambauer et al 2017 \cite{selu} which showed that it was possible to significantly improve on RELUs by giving activation functions ability to control the variance of the underlying signal, and to-date SELU remains the activation function to beat in terms of fast learning rates.

So what constitutes an ideal activation function? There is no clear consensus in the field, but apart from nonlinearity and monotonicity \cite{monotonic}, some of the theoretically desirable properties are $C^{\infty}$ differentiability to help with training, unlimited range to facilitate approximation, and approximating identity at origin to eliminate depenence on the initial guess of  network weights and biases \cite{initialization}. 

Practically speaking, between the sigmoid, RELU and SELU, the three most popular activation functions to date, none satisfy those conditions. All three have limited range. Both RELU and SELU have a non-differentiable kink at the origin. Sigmoids are the only ones that are $C^{\infty}$ differentiable, but they are by far the worst performers due to their other flaws.

In this study, we look at a class of simple $C^{\infty}$ differentiable functions that do approximate identity everywhere, and have infinite range. The functions copy the activations from the previous layer, and then apply a localised nonlinearity with the location, scale and direction of the nonlinearity depending on the training parameters. They are simple one-step reccurrent networks with a localised activation. Such networks do indeed perform as well as RELU and SELU networks in numerical testing on standard examples, but the transformation of the data through the layers is much more tractable than with any other activation functions currently in use.

The paper is organised as follows. Section 2 introduces the Goldilocks activations and fixes the notation. Section 3 is divided in two subsections, one with the  geometrical intuition behind Goldilocks activation, and the other with  mathematical justification in three lemmas. Section 4 contains the numerical results, and Section 5 has the conclusions and the discussion.


 

\section{Goldilocks activations}

We focus on activation functions of the form identity mapping plus a nonlinear perturbation, $A(x) = x + g(x)$ where the nonlinearity $g(x)$  is local in nature. We generate the activation function from hump functions such as the Lorentzian  hump
\begin{equation}
f(x) = \frac{1}{\pi}\frac{1}{1+x^2} \label{goldi}
\end{equation}
and the Gaussian  hump
\begin{equation}
f(x) = \frac{1}{\sqrt{2\pi}} e^{-x^{2}/2} \label{gauss}
\end{equation}
and we differentiate between {\it unbiased Goldilocks} with activation of the form
\begin{equation}
A(x) = x + x f(x) \label{unbiased}
\end{equation}
and {\it biased Goldilocks} with activation of the form
\begin{equation}
A(x) = x + f(x) \label{unbiased}
\end{equation}
where $f(x)$ is any symmetric hump function with decaying tails. As the names suggest,  biased activaton generates a nonzero mean contribution to the data, while unbiased activations have zero mean contribution.

The Goldilocks hump functions can be any simple symmetric,  differentiable localised hump functions with fast decaying tails at infinity. We generally prefer the Lorentzian hump  due to its lower computational overhead (only requiring addition and multiplication as opposed to exponentials) and slower decay in the tails (which is helpful in terms of vanishing gradients); it also has the added benefit of being integrable in terms of elementary functions, which is useful when it comes to designing faster training methods. However, the results apply equally to the Gaussian hump, as well as to many other functions of a similar nature. In further text we denote the local nonlinearity by $g(x)$, where $g(x)=f(x)$ in the biased case, and $g(x)=xf(x)$ in the unbiased case.

\begin{figure}[p]
\includegraphics[width=8cm]{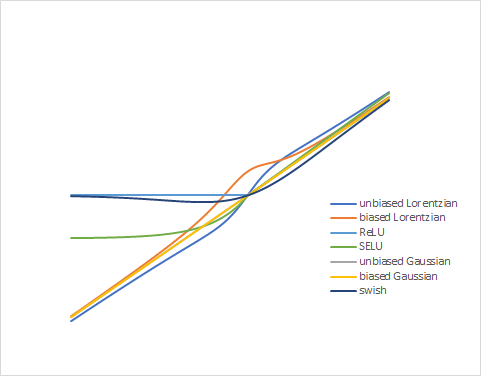}

\caption{Illustration of various activation functions. Biased and unbiased Lorentzian and Gaussian are compared to ReLU,  SELU and swish.}
\label{fig:activations}
\end{figure}

The neural network feed-forward equation can be written  in the {\it direct formulation} as
\begin{equation}
{\bf x}_{n+1} = {\bf W}_{n} {\bf x}_{n} +{\bf b}_{n} +   g\left( {\bf W}_{n} {\bf x}_{n} + {\bf b}_{n} \right) \label{fff}
\end{equation}
or, in the {\it interpretable} formulation,
\begin{equation}
{\bf y}_{n+1} =  {\bf y }_{n} +  {\bf V}_{n}^{-1}  g\left( {\bf  V}_{n} {\bf y}_{n} + {\bf c}_{n} \right) \label{fff1}
\end{equation}
for a sequence of weights matrices ${\bf W}_{n}$ or ${\bf V}_{n}$,  and bias vectors ${\bf b}_{n}$ or  ${\bf c}_{n}$, respectively. 
Intuitively, the first term copies the activations from the previous layer, and the hump function then applies a localised nonlinearity in the current layer. The overall action of the inner layers of the network is a superposition of such local nonlinearities.

The inverse ${\bf V}_{n}^{-1}$ in (\ref{fff1}) is generally the Moore-Penrose pseudoinverse, and when the dimensions of the $n$th and the $(n+1)$st layer coincide, it is almost always the true matrix inverse.

The direct and interpretable formulations (\ref{fff}) and (\ref{fff1}) are connected through
$$ {\bf V}_{n} = {\bf W}_{n} ... {\bf W}_{0}$$
$$ {\bf c}_{n} = {\bf W}_{n}{\bf c}_{n-1} + {\bf b}_{n}$$
$$ {\bf c}_{0} = {\bf b}_{0}$$
\begin{equation}
{\bf x }_{n} =   {\bf V}_{n-1}{\bf y}_{n} + {\bf c}_{n-1}
\label{transform}
\end{equation}

The difference between the direct and interpretable formulations (\ref{fff}) and (\ref{fff1}) is in the vector space that the data is transformed to;  (\ref{fff}) generally allows  the dimension of the $(n+1)$st layer do be different from the $n$th layer, while (\ref{fff1}) projects it back to the domain of the $n$th layer. We generally use the formulation (\ref{fff1}) for theoretical purposes, while (\ref{fff}) is generally more useful computationally.

The backpropagation equation for ${\bf \delta}_{n}$, the sensitivity of the objective function to the activation in $n$th layer, is
\begin{equation}
{\bf \delta}_{n-1} =   {\bf W}_{n} {\bf \delta}_{n} + {\bf W}_{n} g'\left(  {\bf W}_{n} {\bf x}_{n} + {\bf b}_{n} \right)  {\bf \delta}_{n} \label{bpf}
\end{equation}
for the direct formulation (\ref{fff}), and
\begin{equation}
{\bf \delta}_{n-1} =    {\bf \delta}_{n} + g'\left(  {\bf V}_{n} {\bf y}_{n} + {\bf c}_{n} \right) {\bf \delta}_{n} \label{bpf1}
\end{equation}
for the interpretable formulation (\ref{fff1}).

In the interpretable formulation (\ref{fff1}) , it is helpful to think in terms of signal modification between the $n$th and $(n+1)$st layer, $\triangle {\bf y}_{n+1} = {\bf y}_{n+1} - {\bf y}_{n} $,  $\triangle {\bf \delta}_{n+1} = {\bf \delta}_{n+1} - {\bf \delta}_{n} $,which makes the feed-forward equation 
\begin{equation}
\triangle {\bf y}_{n+1} =   {\bf V}_{n}^{-1} g\left({\bf V}_{n}   {\bf y}_{n} + {\bf c}_{n} \right) \label{ffg}
\end{equation}
and the backpropagation equation
\begin{equation}
\triangle {\bf \delta}_{n} =    - g'\left( {\bf V}_{n}   {\bf y}_{n} + {\bf c}_{n} \right)   {\bf \delta}_{n}. \label{bpg}
\end{equation}

The formulation (\ref{ffg}) explains the name "Goldilocks". Namely, if the signal at the $n$th layer is either too large or too small, it passes to the $(n+1)$st layer unmodified to the leading order; the signal is only modified going from the $n$th to the $(n+1)$st layer if it is "just right" in magnitude.

Those are the key properties of any Goldilocks network: the signal is preserved through the layers, but the network can be trained to locally deform the signal non-linearly as required. This is a significant difference to most of the standard activation functions; RELU- and SELU-like activations introduce their nonlinearities by cutting off the lower part of the signal, sigmoid-like activations squeeze the signal into a band, leaky RELUs preserve the entire signal but deform it in a half-plane, and soft exponentials deform it everywhere.

As we demonstrate, both properties are what gives Goldilocks activation their strength. Signal preservation allows the nonlinearity to be inserted anywhere without generating excessive penalties, and the locality of the deformation limits the number of neurons required for preserving the information content of the signal, so they are free to be employed elsewhere.

\section{Properties of Goldilocks activations}

\subsection{Geometrical  Interpretation}

\begin{figure}[p]

\includegraphics[width=8cm]{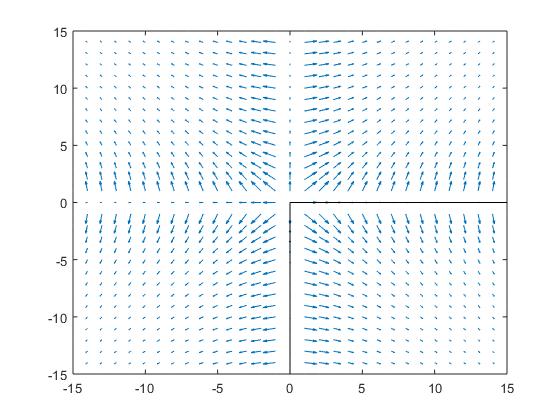}

\includegraphics[width=8cm]{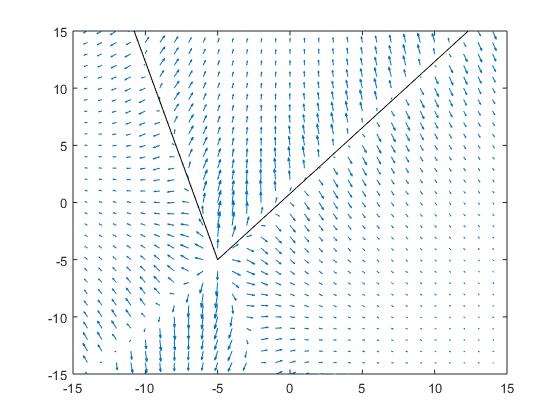}

\caption{Phase diagram illustrating the action of a two -neuron Goldilocks layer in the direct formulation (\ref{fff}) (a), and the interpretable formulation (\ref{fff1}) (b). Arrows start at regularly spaced grid points in the $n$th layer, and they point to their respective positions in the $n+1$st layer. Separating hyperplanes are show in black. The Goldilocks function is unbiased Lorentzian.}
\label{fig:geometry}
\end{figure}

To understand the action of the Goldilocks network, we focus on the $i$th neuron in the $n$th layer of the feed-forward equation (\ref{ffg}).

Taking the superscript $(i)$ to denote the $i$th row,  from the equation (\ref{ffg}), $g$ is constant along a linear hyperplane ${\bf V}_{n}^{(i)}  {\bf y}_{n} + {\bf c}_{n}^{(i)} = 0$, and it only varies in the direction perpendicular to this hyperplane.

The Goldilocks function $g$ is non-zero in the vicinity of the origin and it decays rapidly in both directions away from the origin. In other words, each neuron is selecting a separating hyperplane, and moving data points in the vicinity of this hyperplane smoothly by a fixed distance in the direction normal of the hyperplane. Points sufficiently far from the hyperplane are unaffected. The effective radius of the distance from the hyperplane within which points are affected  is controlled by the norm of the vector ${\bf V}_{n}^{(i)}$. A single hyperplane can then be understood to act as a "broom", sweeping all points in the vicinity of the hyperplane in the perpendicular direction by a fixed distance, as shown in Figure \ref{fig:geometry}(a).

The collection of hyperplanes in each layer, ${\bf V}_{n}^{(i)} {\bf y}_{n} + {\bf c}_{n}^{(i)} = 0\ \forall i$, envelopes a convex cone along whose boundary the points are transformed.
The vertex of the cone and the focus of the motion is at $- {\bf V}_{n}^{-1} {\bf c}_{n} $.

The resulting cone acts as a funnel through the vertex. Points around the edges and the vertex of the cone are swept towards the interior of the cone n, as shown in Figure \ref{fig:geometry}(b). The points sufficiently far from the edges are unaffected.

The next layer  generally produces a different vertex and a different cone, which performs the next step in adjusting the points. Such a sqeuence of cones can result in a flow of arbitrary complexity, limited only by the depth of the network.

Component-wise application of $g$ generates movement across the axes of the domain of $g$; the drift ${\bf c}_{n}$ allows us to set the focus of this movement

\subsection{Statistical properties}

In this section we cover the statistics of data as it passes through Goldilocks layer.

\begin{theorem}[Moments  lemma for a Layer]
\label{momentsm}
Let ${\bf X} \in \mathbb{R}^{m}$ be a normal random variable with mean $\mu$ and the covariance matrix ${\bf \Sigma}$, let $A(x) = x+g(x)$ be the Goldilocks activation and let $u({\bf x}) = A({\bf W}{\bf x}+{\bf b})$, and ${\bf W} \in \mathbb{R}^{n\times n}$ and ${\bf b} \in \mathbb{R}^{m}$.

Then the transformed random variable ${\bf Y}=u({\bf X})$ has the mean $\mu_{{\bf Y}}$ and the covariance matrix ${\bf \Sigma}_{\bf Y}$ given approximately as
$$\begin{array}{cl}   \mu_{\bf Y}  \approx & {\bf W}\mu + {\bf b} + g({\bf W}\mu + {\bf b})  +  \\
& +  \frac{1}{2} \left[ g''({\bf W}\mu + {\bf b})^{(i)} \sum_{j,k} W_{ij} W_{ik} \Sigma_{jk} \right]_{i} \end{array}$$
$$\begin{array}{cl} {\bf \Sigma}_{\bf Y} \approx   &   {\bf W}^{\tau}{\bf \Sigma}{\bf W}  + g'({\bf W}\mu + {\bf b}) \left[ {\bf W \Sigma} + \left( {\bf W \Sigma} \right)^{\tau} \right] + \\
& +  g'({\bf W}\mu + {\bf b})^{2} {\bf W \Sigma W}^{\tau} + \\
&  \frac{1}{2} \left[  g''({\bf W}^{(i)}\mu + b_{i}) g''({\bf W}^{(j)}\mu + b_{j}) U_{i}U_{j} \right]_{ij} \end{array}$$
where 
$$U_{i} = \sum_{k,l} W_{ik}W_{il}\Sigma_{kl}$$
and the superscript $(i)$ denotes the $i$th row.
\end{theorem}

We do not give the proof of the Moments Lemma. It is obtained by straightforwardly plugging in $f({\bf x})$ in the unscented formula of Hendeby and Gustaffson \cite{moments}.

The expressions in the Moments Lemma do appear reasonably complicated, due to the fact that they incorporate mixing terms arising from the interaction of all neurons in a layer. It becomes much clearer if we restrict it to a single neuron, where the target function becomes one dimensional.

\begin{corollary}[Moments  lemma for a Neuron]
\label{moments1}
Let $X$ be a random variable with mean $\mu$ and variance $\sigma^{2}$, and  let $A(x) = x+g(x)$ be the Goldilocks ativation, and let $u(x) = A(wx + b)$ where $w$ and $b$ are real numbers.

Then the transformed random variable $Y=u(X)$ has the mean $\mu_{Y}$ and variance $\sigma_{Y}^{2}$ given approximately as
\[   \mu_{Y} \approx w \mu + b + g(w\mu + b) + \frac{1}{2} \sigma^{2}w^{2} g''(w\mu+b) , \]
\[ \begin{array}{cl}  \sigma_{Y}^{2} \approx & \sigma^{2} w^{2}+ 2 \sigma^{2}w^{2} g'(w\mu+b) + \sigma^{2}w^{2} g'(w\mu+b)^{2} + \\
& +\frac{1}{2} \sigma^{4}w^{4} g''(w\mu+b)^{2}. \end{array} \]
\end{corollary}

In particuar, for unbiased goldilocks $g({\bf 0}) = g''({\bf 0}) = 0$, $g'({\bf 0}) = a {\bf I}$, while for biased goldilocks $g({\bf 0}) = b {\bf I}$, $g'({\bf 0}) = {\bf 0}$, $g''({\bf 0}) =  c {\bf I}$ for some constants $a,b,c$. Hence, for unbiased Goldilocks, the nonlinearity $g(x)$ does not contribute any bias to the transformed variable, but it does contribute to the variance. Conversely, the nonlinearity of biased Goldilocks contributes both bias and variance.

More generally, for all Goldilocks functions, $g'$ and $g''$ take both positive and negative values. The Moments lemma therefore implies that  each layer can  be trained to independently increase or decrease the mean and the variance of the incoming signal, or leave them unchanged. It only modifies signals with mean and variance in a specific range, and it leaves other signals unchanged. 

This is at least as powerful as the "self-normalising" property of SELU networks of Klambauer et al \cite{selu}, and in many ways it is much stronger. For example, a single SELU layer would struggle to leave a signal unchanged, and it can not select the signal by mean and variance on the level of a single neuron.

Incidentally, there is no equivalent of the Moments lemma for any of the standard activation functions such as RELU, SELU, sigmoid or others. The moments lemma crucially depends on the fact that the transformation function $f$ is almost linear.

\subsection{Goldilocks network as ODE}

In this section, we move away from discrete internal layers, and develop the theory of a single continuous inner layer for Goldilocks networks. 

The starting point is equations (\ref{ffg}) and (\ref{bpg}).  They describe the transfer of information from the $n$th to $(n+1)$st discrete layer. Instead of letting $n$ only take discrete values corresponding to discrete inner layers, we can let $n$ vary continuously, corresponding to a continuum of layers. To that end, we note that $\triangle(n+1)$ = $(n+1)-n=1$, so
$$\triangle {\bf y}_{n+1} = \frac{\triangle {\bf y}_{n+1}}{\triangle (n+1)} \approx \frac{d {\bf y}_{n}}{d n}.$$
Extending the domain of $n$ to real numbers, we pass to the continous version of (\ref{ffg}), (\ref{bpg}), 
\begin{equation}
\frac{d}{d n} {\bf y}_{n} = {\bf V}_{n}^{-1}  g\left( {\bf V}_{n} {\bf y}_{n} + {\bf c}_{n} \right), \label{ffc}
\end{equation}
\begin{equation}
\frac{d}{d n} {\bf \delta}_{n} =   - g'\left( {\bf V}_{n} {\bf y}_{n} + {\bf c}_{n} \right)  {\bf \delta}_{n}. \label{bpc}
\end{equation}

The backpropagation equation (\ref{bpc}) can be integrated explicitly to give
\begin{equation}
{\bf \delta}_{n} =  \exp \left[  -\int  {\bf V}_{n} ^{-1}  g\left( {\bf V}_{n} {\bf x}_{n} + {\bf c}_{n} \right)  dn\right] {\bf \delta}_{N}. \label{bps}
\end{equation}

The network equation  (\ref{ffc}) is a bit more difficult in the general case, but there is a specific case wich is quite simple, namely the limit of slowly varying weights and biases,
$$ \left| \frac{d}{d n} {\bf V}_{n}  \right| +  \left| \frac{d}{d n} {\bf c}_{n}  \right| <<  \left| \frac{d}{d n} {\bf x}_{n}\right|. $$

In this limit, we can write (\ref{ffc}) as
$$
\frac{d}{d n} \left( {\bf V}_{n} {\bf y}_{n} + {\bf c}_{n} \right) =   g\left( {\bf V}_{n} {\bf y}_{n} + {\bf c}_{n} \right)
$$

 For Lorentzian Goldilocks, this is 
$$ \frac{d}{d n} \left( {\bf V}_{n} {\bf y}_{n} + {\bf c}_{n} \right) = \frac{ 1}{\pi}\frac{{\bf V}_{n} {\bf y}_{n} + {\bf c}_{n} }{1+ \left( {\bf V}_{n} {\bf y}_{n} + {\bf c}_{n} \right)^{2}},$$
and for Gaussian goldilocks it is 
$$ \frac{d}{d n} \left( {\bf V}_{n} {\bf y}_{n} + {\bf c}_{n} \right) = \frac{1}{\pi} \left( {\bf V}_{n} {\bf y}_{n} + {\bf c}_{n} \right) \exp \left(  {\bf V}_{n} {\bf y}_{n} + {\bf c}_{n} \right).$$

Defining the $n$th layer activation as 
\begin{equation} {\bf A}_{n} = {\bf V}_{n} {\bf y}_{n} + {\bf c}_{n} \label{activation} \end{equation}
we can separate the variables component-wise to get
\begin{equation}\frac{1+ {\bf A}_{n}^{2} }{{\bf A}_{n}} d{\bf A}_{n} = \frac{1}{\pi}  {\bf 1} dn \label{odediff}  \end{equation}
in the Lorentzian case, and
\begin{equation}\frac{\exp {\bf A}_{n}^{2} }{{\bf A}_{n}} d{\bf A}_{n} = \frac{1}{\pi}  {\bf 1} dn \label{odediffG}  \end{equation}
where ${\bf 1}$ denotes a vector composed of 1s. 

Integrating and substituting the initial condition at $n=0$, this becomes
\begin{equation}
 \frac{1}{\pi}  {\bf 1} n = {\bf C}_{n} + \ln {\bf A}_{n}^{2} + {\bf A}_{n}^{2} \label{solnL}
\end{equation}
for the Lorentzian, and
\begin{equation}
 \frac{1}{\pi}  {\bf 1} n = {\bf C}_{n} + Ei \left( {\bf A}_{n}^{2} \right) =  {\bf C}_{n} + \ln {\bf A}_{n}^{2} + {\bf A}_{n}^{2} + \frac{1}{4} {\bf A}_{n}^{4} + ... \label{solnG}
\end{equation}

Equations (\ref{solnL}), (\ref{solnG}) do not have closed formula inverses. However both define monotonically increasing, invertible functions of the rows of ${\bf A}_{n}$.

We can therefore state the main result of this section, namely

\begin{theorem}[Invertibility]
\label{invert}
A Goldilocks neural network  (\ref{fff}) or (\ref{fff1}) is invertible if and only if each ${\bf W}_{n}$ has full rank.
Inverse propagation equations in the direct and interpretable formulations are given by
\begin{equation}
{\bf x}_{n-1} = {\bf W}_{n} {\bf x}_{n} +{\bf b}_{n} -   g\left( {\bf W}_{n} {\bf x}_{n} + {\bf b}_{n} \right) \label{finv}
\end{equation}
\begin{equation}
{\bf y}_{n-1} =  {\bf y }_{n} -  {\bf V}_{n}^{-1}  g\left( {\bf  V}_{n} {\bf y}_{n} + {\bf c}_{n} \right) \label{finv1}
\end{equation}
\end{theorem}

The inversion formulas (\ref{finv}) and (\ref{finv1}) are direct consequences of the ODE formulation, and they are obtained by  inversion of the $n$-dependence in (\ref{ffc}). These inversion formulas are significantly simpler than usual inversion formulas in non-residual neural networs, which typically require inversion of the activation function.

In the real world, the restriction that each  ${\bf W}_{n}$ must have full rank is not difficult to fulfill, since the set of tensors with non-full rank has Lebesgue measure zero. Rank reduction is therefore almost always caused by network architecture, typically either by pooling layers or by explicit dimensionality reduction of specific layers.

Apart from proving the invertibility theorem, the ODE formulation is primarily helpful when it comes to determining the required number of layers. While the closed forms (\ref{solnL}), (\ref{solnG})  have theoretical uses, in practice equations (\ref{ffc}), (\ref{bpc}) are integrated numerically.

\section{Numerical results}

In this section, we look at the numerical performance of deep Goldilocks networks.

\subsection{Toy problem}

We begin by presenting a simple toy problem. The problem is two-phase classification in 2d. 

Initial data was sampled from three independent 2d Gaussian variables, one centered at (-1,0) (blue) and two centered at (1,2) and (1,-2) respectively (amber), with all standard deviations equal to 1 and all correlations equal to zero. 100 points were sampled from each distribution. 

The data is separated via a six-layer biased Lorentzian Goldilocks, with two neurons per layer. The output layer has a single neuron, with linear activation in the first example and sigmoid in the second. The initial guesses for the elements of the weight matrices and bias vectors were sampled from a uniform distribution on $\pm 0.005$. The same initial guess was used in both examples.

\begin{figure}[p]
\centering
\includegraphics[width=4cm]{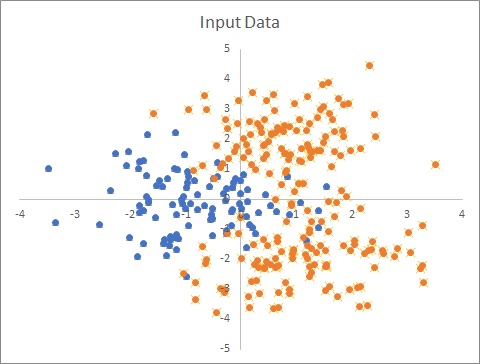}

\includegraphics[width=4cm]{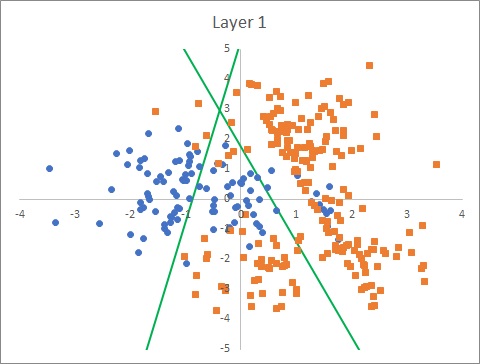}
\includegraphics[width=4cm]{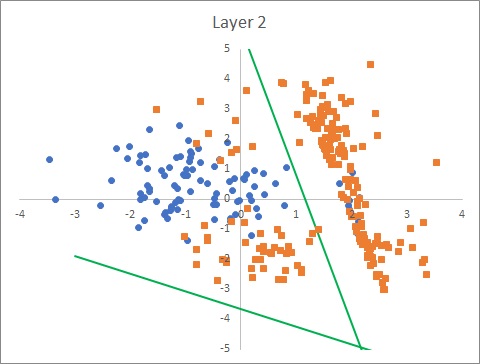}
\includegraphics[width=4cm]{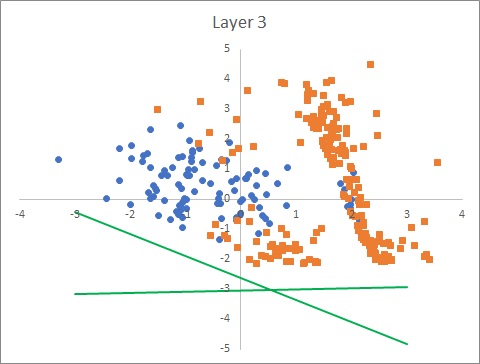}
\includegraphics[width=4cm]{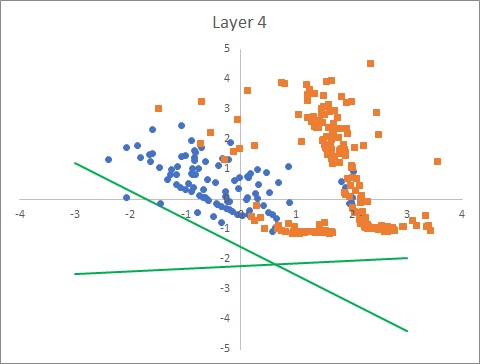}
\includegraphics[width=4cm]{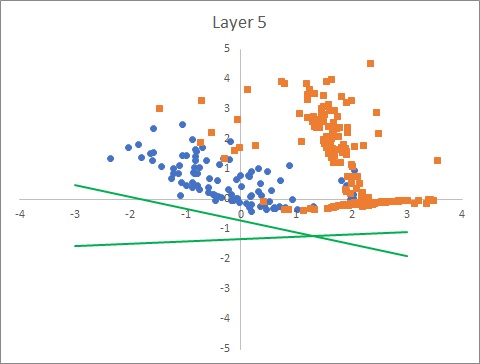}
\includegraphics[width=4cm]{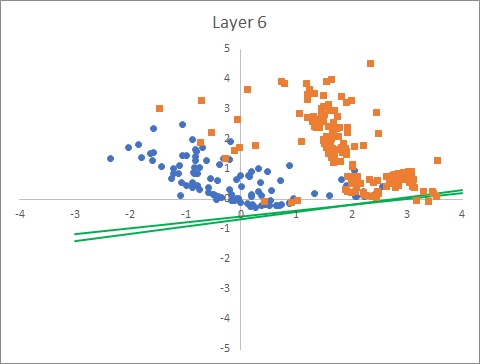}
\includegraphics[width=4cm]{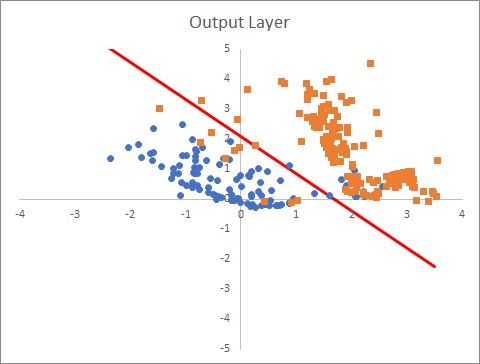}
\centering
\includegraphics[width=0cm]{loo}
\caption{Two-phase classifier in two dimensions with linear output, in interpretable coordinates. Identity mapping as the input layer, six inner layers and a linear output layer. The Goldilocks function is biased Lorentzian, and the initial guess is random. Inner layer separating hyperplanes are shown in green, and the single output layer separating hyperplane is shown in red.}
\label{fig:toy}
\end{figure}

\begin{figure}[p]
\centering
\includegraphics[width=4cm]{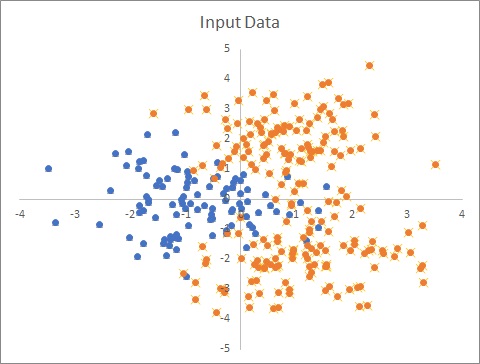}

\includegraphics[width=4cm]{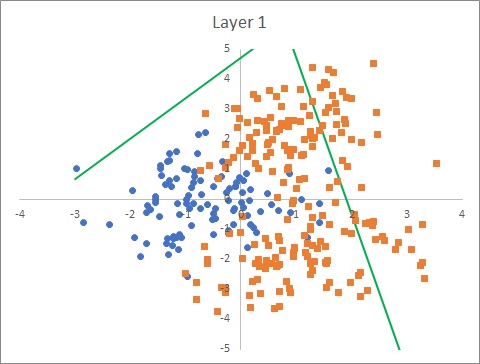}
\includegraphics[width=4cm]{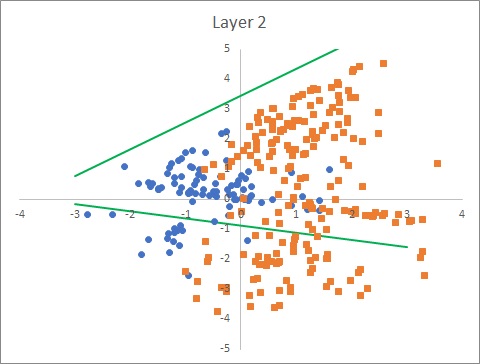}
\includegraphics[width=4cm]{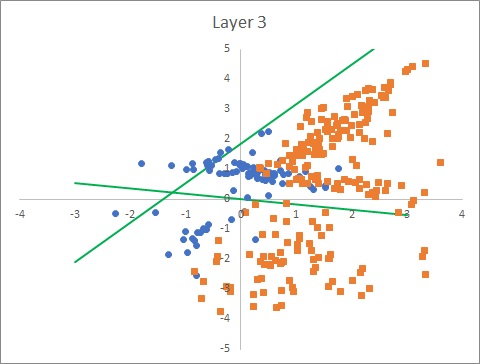}
\includegraphics[width=4cm]{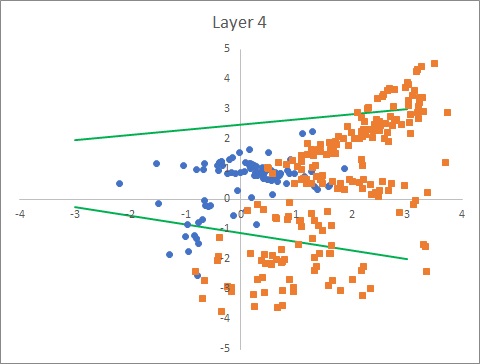}
\includegraphics[width=4cm]{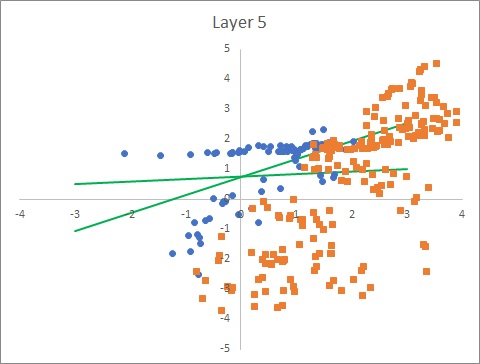}
\includegraphics[width=4cm]{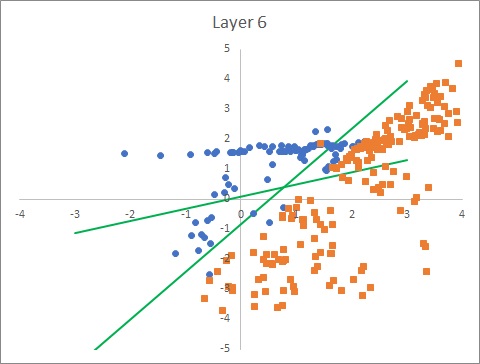}
\includegraphics[width=4cm]{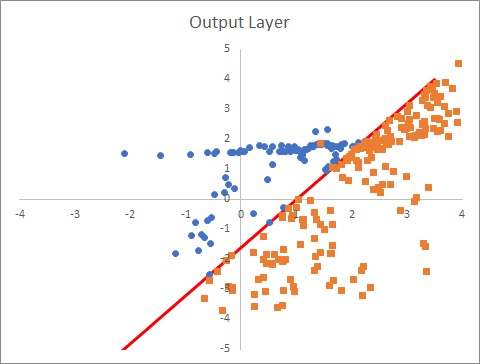}
\centering
\includegraphics[width=0cm]{loo}
\caption{Two-phase classifier in two dimensions with sigmoid output. Everything is identical to Figure \ref{fig:toy}, except the activation function of the output neuron which is sigmoid here, as opposed to linear in Figure  \ref{fig:toy}.}
\label{fig:toy1}
\end{figure}

The results are shown in Figures \ref{fig:toy} and \ref{fig:toy1}.

Qualitatively, we see a three-stage behaviour. 

In the initial, "chopping" stage (Layers 1, 2), the biggest gain is achieved by running separating hyperplanes through a high concentration of data points and moving them by a relatively large distance. The classification achieved in these layers is imprecise and it leaves a significant proportion of data misclassified. However, it directly leads to the subsequent stage.

In the intermediate, "carving" stage (Layers 3, 4), there is seemingly no meaningful progress at further classification. Separating hyperplanes are equally likely to run through a high concentration of data points as through regions with no data points at all, and any movements are fairly minor. The focus in these layers is on deformation of the data.

In the late, "pruning" stage (Layers 5, 6), the separating hyperplanes make no attempt to run through regions of high concentration of data, or, if they do, the modification they make is minimal. Instead, they are fine-tuning the final deformation to pass to the output layer.

The output layer inherits the suitably linearised data, and it draws the final separating hyperplane(s) which classify the data.

This behaviour is qualitatively similar between the linear activation (Figure  \ref{fig:toy}) and the sigmoid activation (Figure  \ref{fig:toy1}) in the output layer. The task is more difficult with the linear activation in the output layer, since  linear output forces the inner layers to group the data into two parallel hyperplanes, and siginificant errors can still be generated when the data is separated, but not properly aligned. Arguably, this is what we see in Layers 5 and 6 in Figure  \ref{fig:toy}, where the data is already separated, but not aligned. Conversely, sigmoid output only generates an error with the choice of the separating hyperplane, and there is no penalty for not aligning the data away from the separating hyperplane. 

Consequently, in our example, the exact same configuration generates a much smaller classification error with the sigmoid output layer compared to the linear output layer (classification error 1.1114\% vs 5.8419\%).

\subsection{CIFAR}

We have compared the Goldilocks network to RELU and SELU on CIFAR-10 and CIFAR-100 example sets.

To avoid selecting the architecture that artificially favours Goldilocks activations, we have used the neural network archiecture from a GitHub tutorial on Self Normalizing Neural Networks which is specifically used to compare RELU and SELU activations on CIFAR data sets (https://github.com/bioinf-jku/SNNs), and modified it to include Goldilocks activations. The network consists of five convolution layers and three max pool layers (convolution / max pool / convolution / max pool / convolution / convolution / convolution / max pool) followed by two fully connected layers. 

We created four copies of the neural network, one for each actvation. Within each copy, the same activation function was used for each inner layer, and the output layer was linear. The activation functions we compared are SELU, RELU, unbiased Lorenzian Goldilocks and biased Lorenzian Goldilocks. The learning rate and, where applicable, regularization and dropout, were the same for all activations.

\begin{figure}[p]

\includegraphics[width=9cm]{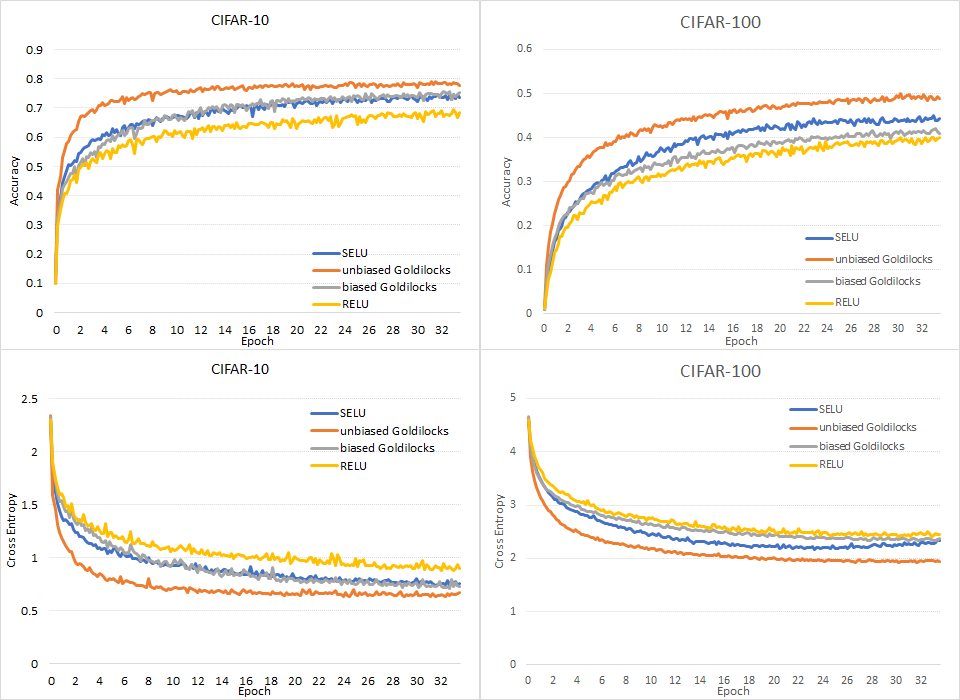}

\caption{Training on CIFAR-10 and CIFAR-100, comparison of test accuracy and test cross entropy between SELU, unbiased Lorentzian Goldilocks, biased Lorentzian Goldilocks and RELU.}
\label{fig:cifar-10}
\end{figure}

In line with  expectations, unbiased activations showed significantly faster learning than biased activations, as shown in Figure \ref{fig:cifar-10}. For both CIFAR-10 ad CIFAR-100, unbiased Goldilocks was significantly the fastest learning method, followed by SELU. Biased Goldilocks and RELU were consistently close to each other. 

Without regularization or dropout, unbiased Goldilocks reached peak test accuracy in as few as 2 or 3 epochs, compared to 7 or 8 epochs for SELU. However, Goldilocks then started to overtrain, and its peak test accuracy was lower than that of SELU, but still significantly higher than those of either biased Goldilocks or RELU.

This suggests that, neuron-per-neuron, biased Goldilocks has more effective degrees of freedom than SELU, SELU has more than either biased Goldilocks or RELU, while biased Goldilocks or RELU are roughly similar.

To test this, we repeated the training with L2 regularization of the loss function $L$,
$$
L_{\beta}  = L  + \beta \sum_{n} | {\bf W}_{n} |^{2} .
$$

 As expected, reasonable values of $\beta$ ($\beta \sim 0.01$) significantly improved the performance of Goldilocks, making it not overtrain in 20 epochs. Peak test accuracy of regularized unbiased Goldilocks  surpassed the best result for SELU (CIFAR-10: unbiased Goldilocks peak test accuracy 77.55\%, with $\beta = 0.02$; SELU peak test accuracy  77.11\%  with $\beta=0.01$; CIFAR-100: unbiased Goldilocks peak test accuracy 49.91\%, with $\beta = 0.01$; SELU peak test accuracy  44.58\%  with $\beta=0$). For $\beta > 0.02$, all networks deteriorated in performance, but the deterioration was least pronounced for unbiased Goldilocks, and the gap between biased Goldilocks and other activations increased with higher $\beta$.

We also trained the networks with dropout after the last fully connected layer and no regularization. This generated more mixed results. Reasonable levels of dropout (drop probability $<$ 50\%) had no effect on any of the activations, with all activations reaching roughly the same peak test accuracies and beginning overtraining at roughly the same epochs as with no dropout. Significantly high levels of dropout (drop probability $\sim$  95\%) slowed the learning rates significantly over all networks, and the effect was particularly pronounced on unbiased Goldilocks.

Why did Goldilocks suffer disproportionately from dropout?  Geometrically, dropping a single coordinate projects the data points to some, possibly distant, hyperplane. This is qualitatively similar to the result of a single RELU activation, which also projects a number of data points to a possibly distant hyperplane. It is also not very dis-similar from the result of a single SELU activation, where the projection is smoothed out over a region of non-zero thickness. For RELU and SELU, the noise generated by dropout is therefore at most comparable to the signal of the trained activation functions.

On the other hand, Goldilocks acts by slowly moving data points from one layer to another. A sudden projection of a data point to a possibly distant hypersurface can be significantly larger than the regular action of the activation function, and the dropout noise to signal ratio is consequently much higher for Goldilocks than for either RELU or SELU.

Conversely, because SELU and RELU routinely project data points to distant hypersurfaces, they are disproportionately affected by small changes to the trajectory of such projections, such as those generated by L2 regularization.

In short, activation functions with 'smooth' trajectores such as Goldilocks react well to 'smooth' regularizations such as the L2 penalty, while activations with  'jerky' trajectories such as RELU or SELU react well to 'jerky' regularizations such as dropout.

Consequently, it will be difficult to obtain more detailed neuron-per-neuron comparison between such activations in any situation requiring regularization.

\subsection{Interpretability}

\begin{figure}[p]

\includegraphics[width=10cm]{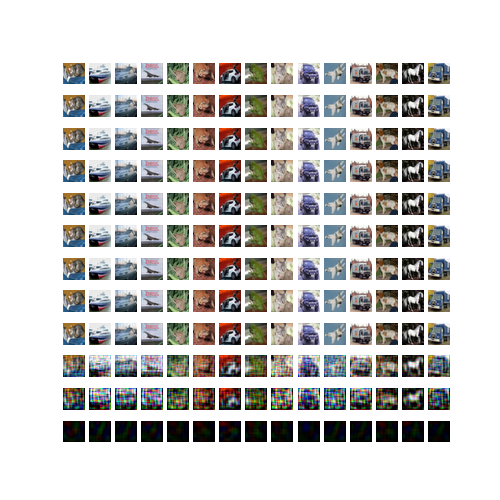}

\caption{First fifteen images from the test set of CIFAR-10, deformation through the layers as projected to the input layer, with linear activation. The network was trained with the biased Goldilock activation. The input layer 0 is on top, the output layer 11  is at the bottom.}
\label{fig:interpret1}
\end{figure}

The interpretable neural network formulation (\ref{fff1}) projects each layer back to the coordinates of the input layer. This enables a direct comparison of how each data point is deformed in each layer.

\begin{figure}[p]
\includegraphics[width=10cm]{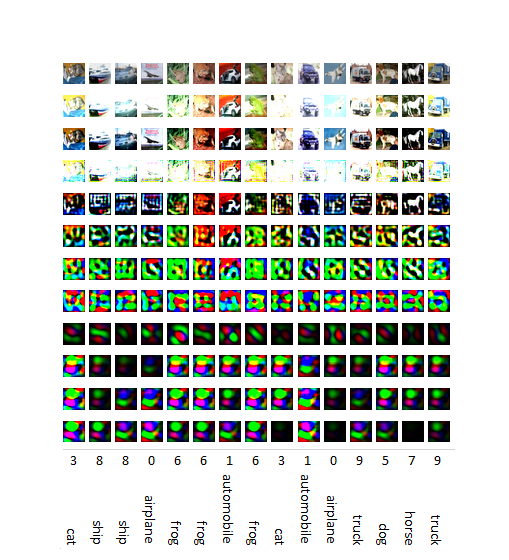}
\caption{First fifteen images from the test set of CIFAR-10, deformation through the layers as projected to the input layer, with biased Goldlocks activation. The input layer 0 is on top, the output layer 11  is at the bottom. Results of the classificaton are provided below the images.}
\label{fig:interpret2}
\end{figure}

\begin{figure}
\includegraphics[width=10cm]{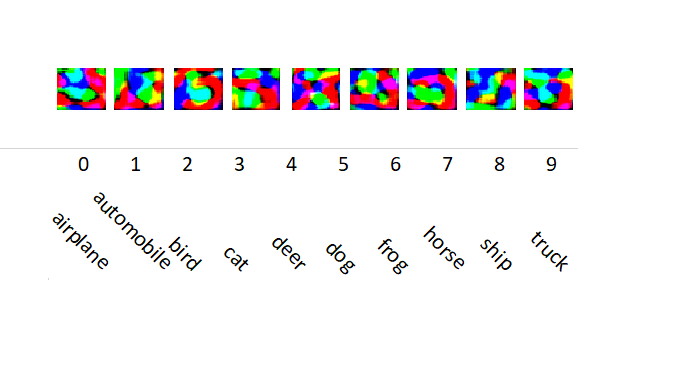}
\caption{Backprojection of CIFAR-10 categories.}
\label{fig:interpret3}
\end{figure}

In low dimensional systems, this can be visualised on the entire data set as in Figures \ref{fig:toy} and \ref{fig:toy1}. For higher dimensional data sets such as CIFAR-10 or CIFAR-100, the visualisation is done on the level of individual data points, as shown in Figures \ref{fig:interpret1},  \ref{fig:interpret2},  \ref{fig:interpret3}.

Figure  \ref{fig:interpret1} shows the backprojection of layer states when there are no nonliearities in the activation function. In this situation, the only change to  CIFAR-10 images comes from loss of dimensionality in pooling layers 2, 4, 8, and the bottleneck layers 9 and 10.  

When the full activation is retained, as in Figure  \ref{fig:interpret2}, the deformation of individual layers is more pronounced. Individual images retain some basic reseblace to the input layer through first fve layers or so, but then they become more difficult to interpret. 

In the output layer, the figures have been reduced to superpositions of simple category images. The pure category images can be isolated by backprojecting vectors $(1,0,0,....), (0,1,0,...) ... (..,0,0,1)$. For this particular example, they are shown in Figure  \ref{fig:interpret3}.

\section{Conclusion}

The class of neural networks studied in this paper represents a  new family of neural networks with strong interpolation properties and faster learning rates than most, or even any other neural network architectures presently in use. The general feature of such networks is that each layer creates an identical copy of the previous layer, and then applies to it a localised nonlinear Goldilocks transformation. The Goldilocks transformation could be the  Lorenzian, the Gaussian, or any of a number of similar functions. 

The networks are similar in architecture to the Neural ODEs of Chen et al \cite{ode}, and they can be immediately restated in ODE terms by making the layer index $n$ continuous and changing the finite difference operator $\triangle$ in (\ref{ffg}) and (\ref{bpg}) to a continous derivative. 

In standard examples such as CIFAR-10 and CIFAR-100, Goldilocks networks perform  better than, or, at worst, comparably to RELU and SELU networks.

The true strength of Goldilocks, however, is probably in reversibility and interpretability.  In the interpretable formulation, local deformations of a Goldilocks layer are globally reversible, and the trajectory from the input to the output layer is continuous. We therefore have a contiguous link through all layers of the network, making it straightforward to interpret the action of each layer.

All examples so far are fairly simple, and there is a signifcant amount of further work to be done on testing the true power of Goldilocks in real world learning scenarios. 

Nonetheless, Goldilocks activations are an interesting new class of activation functions whose features include interpretability, reversibility, intuitively clear geometrical interpretation, and very strong performance on standard samples.


%





\section*{Acknowledgment}

The authors would like to acknowledge the support of the Wellcome Trust grant 103952/Z/14/Z.


\begin{thebibliography}{1}


\bibitem{cybenko} G Cybenko  (1989) \emph{Approximation by Superpositions of a Sigmoidal Function} Math. Control Signals Systems 2:303-314.
\bibitem{elliott} DL Elliot \emph{A better activation function for artificial neural networks}, ISR Technical Report TR 93-8, University of Maryland, College Park, MD 20742., CiteSeerX 10.1.1.46.7204
\bibitem{quadratic} J Bergstra, G Desjardins, P Lamblin, Y Bengio (2009) \emph{Quadratic polynomials learn better image features} Technical Report 1337. Département d’Informatique et de Recherche Opérationnelle, Université de Montréal.
\bibitem{training} X Glorot, Y Bengio (2010) \emph{Understanding the difficulty of training deep feedforward neural networks} International Conference on Artificial Intelligence and Statistics (AISTATS'10), Society for Artificial Intelligence and Statistics
\bibitem{sqnl} A Wuraola, N Patel (2018) \emph{SQNL:A New Computationally Efficient Activation Function}, 2018 International Joint Conference on Neural Networks (IJCNN), Rio Rio de Janeiro, Brazil: IEEE, pp. 1–7
\bibitem{phase} MD Klimek, M Perelstein (2018) \emph{Neural Network-Based Approach to Phase Space Integration}. arXiv:1810.11509 [hep-ph].
\bibitem{relu} V Nair, GE Hinton (2010) \emph{Rectified Linear Units Improve Restricted Boltzmann Machines} 27th International Conference on International Conference on Machine Learning, ICML'10, USA: Omnipress, pp. 807–814, ISBN 9781605589077
\bibitem{leaky} AL Maas, AY Hannun, AY Ng(2013) \emph{Rectifier nonlinearities improve neural network acoustic models} Proc. ICML. 30 (1).
\bibitem{isrlu} B Carlile, G Delamarter, P Kinney, A Marti, B Whitney, Brian (2017) \emph{Improving Deep Learning by Inverse Square Root Linear Units (ISRLUs)} arXiv:1710.09967 [cs.LG].
\bibitem{bipolar} L Eidnes, A Nøkland (2018) \emph{Shifting Mean Activation Towards Zero with Bipolar Activation Functions} International Conference on Learning Representations (ICLR) Workshop. arXiv:1709.04054.

\bibitem{prelu} K He, X Zhang, S Ren, J Sun (2015) \emph{Delving Deep into Rectifiers: Surpassing Human-Level Performance on ImageNet Classification}. arXiv:1502.01852 [cs.CV].
\bibitem{rrelu} B Xu, N Wang, T Chen, M Li (2015) \emph{Empirical Evaluation of Rectified Activations in Convolutional Network}. arXiv:1505.00853 [cs.LG].
\bibitem{elu} DA Clevert, T Unterthiner, S Hochreiter (2015) \emph{Fast and Accurate Deep Network Learning by Exponential Linear Units (ELUs)}. arXiv:1511.07289 [cs.LG].
\bibitem{sshape} X Jin, C Xu, J Feng, Y Wei, J Xiong, S Yan (2015) \emph{Deep Learning with S-shaped Rectified Linear Activation Units}. arXiv:1512.07030 [cs.CV].
\bibitem{learning} F Agostinelli,  M Hoffman, P Sadowski,  P Baldi (2014). \emph{Learning Activation Functions to Improve Deep Neural Networks}  arXiv:1412.6830 [cs.NE].
\bibitem{softplus} X Glorot, A Bordes, Y Bengio (2011) \emph{Deep sparse rectifier neural networks} (PDF). International Conference on Artificial Intelligence and Statistics.
\bibitem{gelu} D Hendrycks, K Gimpel, (2016). \emph{Gaussian Error Linear Units (GELUs)} arXiv:1606.08415 [cs.LG].
\bibitem{sigmoidw} S Elfwing, E Uchibe, K Doya,  (2017) \emph{Sigmoid-Weighted Linear Units for Neural Network Function Approximation in Reinforcement Learning} arXiv:1702.03118 [cs.LG].
\bibitem{search} P Ramachandran, B Zoph, QV Le (2017) \emph{Searching for Activation Functions} arXiv:1710.05941 [cs.NE].
\bibitem{continuum} LB Godfrey, MS Gashler (2016) \emph{A continuum among logarithmic, linear, and exponential functions, and its potential to improve generalization in neural networks} 7th International Joint Conference on Knowledge Discovery, Knowledge Engineering and Knowledge Management: KDIR. 1602: 481–486. arXiv:1602.01321. Bibcode:2016arXiv160201321G.

\bibitem{fourier} MS Gashler, SC Ashmore  (2014) \emph{Training Deep Fourier Neural Networks To Fit Time-Series Data}  arXiv:1405.2262 [cs.NE].
\bibitem{ode} RTQ Chen, Y Rubanova, J Bettencourt, D Duvenaud (2018) \emph{Neural Ordinary Differential Equations} 32nd NIPS Proceedings.
\bibitem{selu} G~Klambauer, T ~Unterthiner, A ~Mayr and S ~Hochreiter (2017) \emph{Self-Normalizing Neural Networks}, 31st NIPS Proceedings.
\bibitem{monotonic} H Wu (2009). \emph{Global stability analysis of a general class of discontinuous neural networks with linear growth activation functions}. Information Sciences. 179 (19): 3432–3441. doi:10.1016/j.ins.2009.06.006.
\bibitem{initialization} D ~Susillo and L F ~Abbott (2015) \emph{Random Walk Initialization for Training Very Deep Forward Networks}, ICLR Proceedings.
\bibitem{moments} G Hendeby, F Gustafsson. (2007) \emph{On Nonlinear Transformationsof Stochastic Variables and itsApplication to Nonlinear Filtering} Technical Report LiTH-ISY-R-2828, Proceedings ICASSP 2008

\end{thebibliography}
\end{document}